\documentclass{article}
\usepackage{spconf,amsmath,graphicx,hyperref,epstopdf}
\usepackage{amssymb}
\usepackage{xcolor}
\usepackage{dsfont}
\usepackage{multirow}
\usepackage[numbers,sort&compress]{natbib}
\usepackage{colortbl} 
\usepackage{booktabs}
\usepackage{multirow}
\usepackage{subcaption}
\usepackage{lipsum}
\usepackage{wrapfig}

\title{CPATTA: Conformal Supervision Allocation for Active Test-Time Adaptation}

\name{
Tingyu Shi$^{1}$ \qquad 
Fan Lyu$^{2}$ \qquad
Haihua Zhu$^{1}$ \qquad 
Dadi Wang$^{3}$ \qquad 
Shaoliang Peng$^{1}$\sthanks{Corresponding author. Email: \texttt{slpeng@hnu.edu.cn}} }
\address{
$^{1}$ College of Computer Science and Electronic Engineering, Hunan University\\
$^{2}$ Computer Vision Center, Universitat Autònoma de Barcelona \\
$^{3}$ Department of Computing \& Software, McMaster University}

\begin{document}
\ninept
\maketitle
\begin{abstract}
Active Test-Time Adaptation (ATTA) improves model robustness under domain shift by selectively querying human annotations at deployment, but existing methods use heuristic uncertainty measures and suffer from low data selection efficiency, wasting human annotation budget. We propose Conformal Prediction Active TTA (CPATTA), which first brings principled, conformal uncertainty with coverage-aware online calibration into ATTA. CPATTA employs smoothed conformal scores with a top-$K$ certainty measure, an online weight-update algorithm driven by pseudo coverage, a domain-shift detector that adapts human supervision, and a staged update scheme that balances human-labeled and model-labeled data. Extensive experiments demonstrate that CPATTA consistently outperforms the state-of-the-art ATTA methods by around 5\% in accuracy.

\end{abstract}

\vspace{-10px}
\section{Introduction}
\vspace{-10px}
\label{sec:intro}
Test-Time Adaptation (TTA) updates pretrained models using unlabeled test data to address domain shifts common in autonomous driving \cite{sun2022car1}, medical imaging \cite{kondrateva2021mri1}, and speech recognition \cite{sun2017speech1}. However, without ground-truth supervision, existing TTA methods \cite{wang2020tent, niu2022eata, wang2022cotta, niu2023sar} often achieve suboptimal online and post-adaptation performance. Inspired by Active Learning, Active Test-Time Adaptation (ATTA) \cite{gui2024simatta} selectively acquires human annotations to provide supervised signals and improve robustness.

\begin{figure}[t]
\begin{minipage}[b]{1.0\linewidth}
  \centering
  \centerline{\includegraphics[scale=0.22]{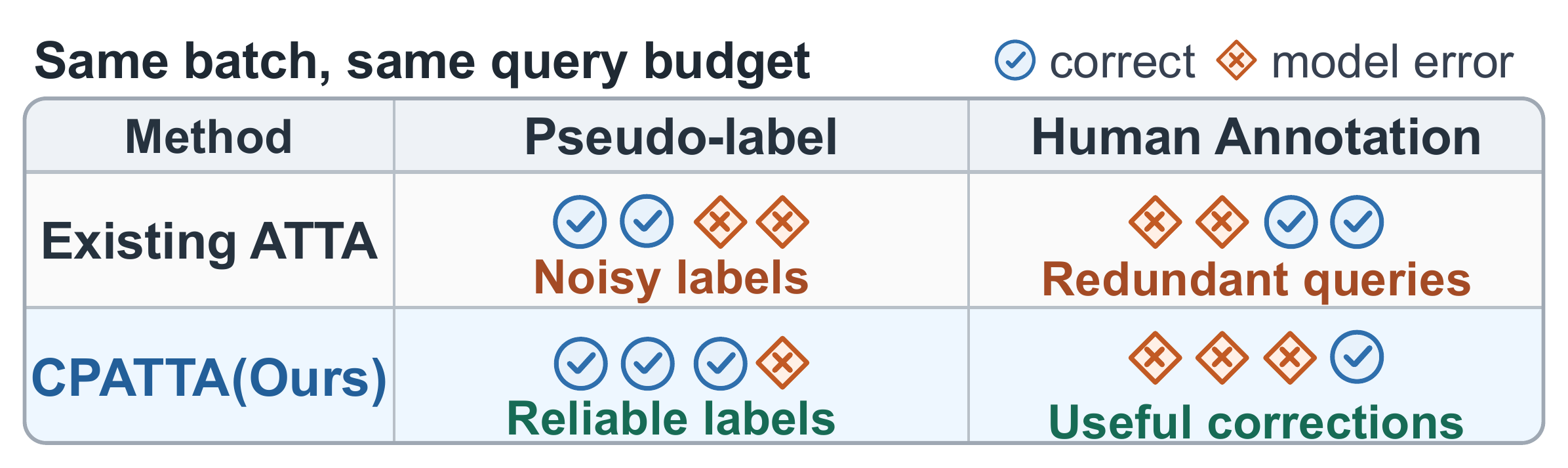}}
 \vspace{-10px}
\end{minipage}
\caption{Comparison between existing ATTA methods and CPATTA. Selection efficiency is defined as the fraction of useful samples: correct predictions for model-annotated (pseudo-label) data and incorrect predictions for human-annotated data. CPATTA achieves higher selection efficiency in both cases, enabling better real-time and post-adaptation performance under the same annotation budget.}
\label{fig:motivation}
\vspace{-15px}
\end{figure}

Existing ATTA methods use different criteria to select samples for human annotation. SimATTA \cite{gui2024simatta} combines fixed entropy thresholds with incremental clustering, while CEMA \cite{chen2024cema} uses a fixed lower threshold and a dynamically adjusted upper threshold. EATTA \cite{wang2025eatta} identifies samples near the boundary between the source and shifted domains by perturbing softmax scores. However, these methods suffer from \textbf{low data selection efficiency}, as shown in Fig. ~\ref{fig:motivation}. Many human annotated samples are redundant because the model can already predict them correctly, leading to \textit{a waste of limited human supervision}. A more effective selection mechanism is therefore needed to maximize annotation utility and adaptation performance.

\begin{figure*}[t] 
  \centering
\includegraphics[scale=0.5]{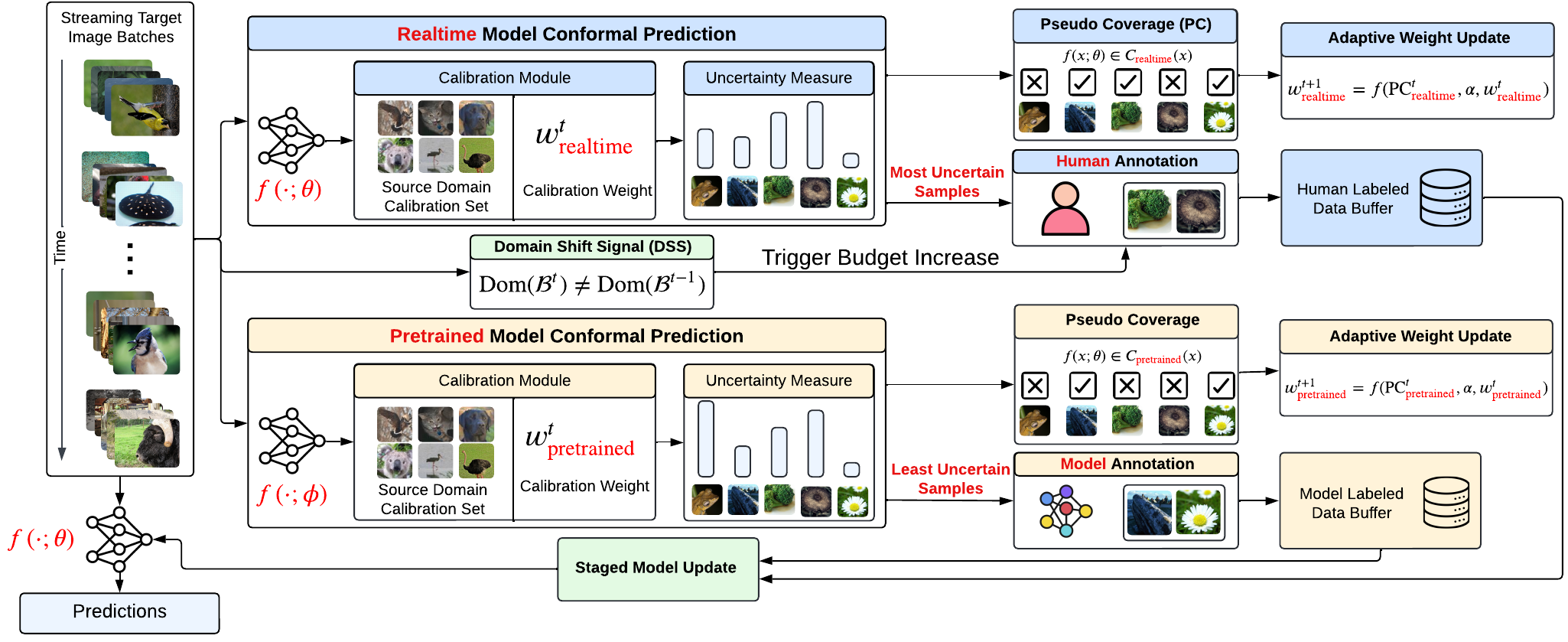}
  \vspace{-10px}
  \caption{\textbf{CPATTA Overview}. The real-time model $f(\cdot;\theta)$ makes the 
  predictions right away when an online batch of data arrives. DSS raises the human budget after detected shifts; real-time and pretrained CPs rank samples for human and model annotation, respectively. Pseudo coverage updates their weights, and the resulting buffers drive staged adaptation.
  }
  \label{fig:schematic}
  \vspace{-15px}
\end{figure*}

To improve ATTA selection, a natural approach is a principled uncertainty measure. \textbf{Conformal prediction (CP)} converts point predictions into prediction sets with coverage guarantees, whose size reflects uncertainty. \textit{However, applying CP to ATTA is non-trivial.} Classical CP assumes exchangeable between calibration data and test data ~\cite{angelopoulos2021gentle, shafer2008cporiginal, angelopoulos2024crc}, violated because source-domain calibration and shifted test samples differ. Recent extensions relax this assumption but cannot guarantee validity under evolving test-time adaptation conditions ~\cite{farinhas2023nexcrc, barber2023nexcptheory, yilmaz2022qtc}. \textit{This gap motivates us to explore how CP can be adapted to improve annotation efficiency in the ATTA setting.}

We propose \textbf{C}onformal \textbf{P}rediction \textbf{A}ctive \textbf{TTA} (CPATTA), which integrates CP into ATTA using coverage-aware uncertainty measures adapted to dynamic test environments. CPATTA has three components. First, smoothed conformal scores and a top-$K$ certainty measure provide fine-grained uncertainty signals for efficient human annotation and reliable pseudo-labeling. Second, an online weight-update algorithm uses pseudo coverage feedback to maintain calibration under domain shifts at a chosen risk level. Third, a domain-shift detector increases human supervision for new domains, preventing early error accumulation. Together, these components enable efficient and reliable annotation, improving both real-time and long-term adaptation. Experiments show that CPATTA outperforms existing ATTA methods.

\vspace{-10px}
\section{Method}
\vspace{-5px}
\label{sec:method}

\subsection{Problem Statement and Conformal Prediction}
\label{sec:ps}
\vspace{-3px}
Let $f(\cdot;\phi)$ be pretrained on source data $\mathcal{D}_S$ and initialized at deployment as $\theta^0=\phi$. For each target batch $\mathcal{B}^t=\{x_i^{S'}\}_{i=1}^{N_t}$, the algorithm selects one subset for human annotation and another for model-generated pseudo-labeling, leaving the remaining samples unlabeled. It then updates the model from $\theta^t$ to $\theta^{t+1}$ using both annotated subsets. ATTA aims to maximize online performance under a limited human annotation budget.

Existing ATTA methods have low data-selection efficiency, making annotation sample selection critical. Conformal Prediction (CP) quantifies model uncertainty using a small labeled calibration set $\mathcal{D}_{cal}=\{(x_i,y_i)\}_{i=1}^n$ and a nonconformity score $\mathcal{S}$. For a test input $x_{n+1}$, its prediction set is
\vspace{-7px}
\begin{equation}
\mathcal{C}(x_{n+1})=\left\{y\in\mathcal{Y}\mid\mathcal{S}(x_{n+1},y)\leq\tau\right\},
\vspace{-5px}
\end{equation}
where
\vspace{-5px}
\begin{equation}
\tau=\text{Quantile}_{ \lceil (n+1)(1-\alpha) \rceil / n }\left(\left\{\mathcal{S}(x_i,y_i)\right\}_{i=1}^n\right).
\end{equation}
For classification, $\mathcal{S}$ is typically one minus the correct-label softmax score. Under exchangeability between calibration and test data, CP guarantees \cite{angelopoulos2021gentle}
$
1-\alpha\leq\mathbb{P}\left(y\in\mathcal{C}(x)\right)\leq1-\alpha+\frac{1}{1+n}.
$
Larger prediction sets indicate greater uncertainty. The source-domain calibration samples are used only to estimate uncertainty, not for training. \textit{In test-time applications where annotation resources are critical, such as autonomous driving with safety risks, or medical imaging with costly expert labeling, this calibration reduces redundant human annotations and improves real-time as well as post-adaptation performance.}
However, this guarantee requires exchangeability, which ATTA violates because source-domain calibration data differ from shifted test batches, creating a \textbf{coverage gap} between target and actual coverage. 

Weighted CP extensions \cite{barber2023nexcptheory, farinhas2023nexcrc} seek to restore coverage through reweighting. Specifically, they give greater influence in the conformal quantile to calibration samples that are more representative of the target distribution, while downweighting mismatched samples. However, this approach requires accurate weights, which are difficult to estimate under dynamic shifts.
Thus, existing CP methods are unsuitable for ATTA, and \textit{this limitation motivates us to develop a coverage-aware adaptive weighting algorithm that uses pseudo-coverage feedback to reduce the empirical coverage gap and improve sample-selection reliability.}


\subsection{Uncertainty-Guided Annotation with CP}
Standard CP relies on prediction set size as an uncertainty proxy, which is too coarse for ATTA. We instead adopt smoothed prediction sets \cite{stutz2021smoothedcp}, converting hard set membership into soft inclusion scores and yielding more fine-grained uncertainty signals that adapt to dynamic test-time shifts.
Given a normalized nonconformity score function $\mathcal{S}$, the soft score of including label $y$ for input $x$ is
\vspace{-5px}
\begin{equation}
E(x, y;\tau) = \sigma \left( \left( \tau - \mathcal{S}(x, y) \right) / \kappa \right),
\label{softscore}
\vspace{-5px}
\end{equation}
where $ \sigma(x) = \frac{1}{1 + e^{-x}}$. $\kappa$ represents temperature. 
For a test input $x$ with label space $\mathcal{Y}=\{1,\dots,L\}$, we sort labels in descending order
\vspace{-5px}
\begin{equation}
E(x,y_{(1)};\tau) \geq E(x,y_{(2)};\tau) \geq \cdots \geq E(x,y_{(L)};\tau),
\label{softsocreorder}
\end{equation}
and define the top-$K$ certainty score as
\vspace{-5px}
\begin{equation}
\text{Cert}_{K}(x;\tau) = \mathbb{E}_{k\in[1,K]}  E(x, y_{(k)}; \tau),
\label{cert}
\vspace{-5px}
\end{equation}
which reflects how strongly the model supports its most plausible labels, which is more informative compared to the hard CP.

Second, we leverage two complementary conformal predictors: one based on the pretrained model $f(\cdot;\phi)$ and one on the real-time adapted model $f\left(\cdot;\theta^t\right)$. For each test batch $\mathcal{B}^t$, these predictors yield thresholds $\tau_{\text{pre}}$ and $\tau_{\text{rt}}$, from which we compute
\vspace{-5px}
\begin{equation}
    \text{Cert}_{K}^{\text{pre}}(x) = \text{Cert}_{K}(x;\tau_{\text{pre}}), \quad
\text{Cert}_{K}^{\text{rt}}(x) = \text{Cert}_{K}(x;\tau_{\text{rt}}).
\vspace{-5px}
\end{equation}
Annotation is then allocated by sending the $N_{\text{human}}$ least-certain samples under the real-time predictor to human annotators, while the $N_{\text{model}}$ most-certain ones under the pretrained predictor receive pseudo-labels. The selected samples are stored in human buffer $\text{Buf}_H$ and model buffer $\text{Buf}_M$ for subsequent parameter updates.

Finally, to adapt annotation effort to domain dynamics, we integrate a domain-change detector. Specifically, we employ the Domain Shift Signal (DSS) \cite{chakrabarty2023dss} to test whether the current batch $\mathcal{B}^t$ originates from a new domain. If a shift is detected, we temporarily increase the human annotation budget from $N_{\text{human}}$ to $N_{\text{human}}' \geq N_{\text{human}}$, reflecting the reduced reliability of prior knowledge and accelerating adaptation at the onset of new domains.


\subsection{Adaptive Weighting for CP in ATTA}
\label{sec:weightupdate}
Although uncertainty-guided annotation improves efficiency, valid coverage is still required for reliable adaptation. Since calibration data come from the source domain, the exchangeability assumption of classical CP is violated under domain shift (Sec.~\ref{sec:ps}). To address this, we introduce a dynamic weighting mechanism that updates CP online during adaptation.

\noindent
\textbf{Pseudo Coverage as Feedback}.
Since ground-truth labels are unavailable at test time, true coverage cannot be computed. We thus define \emph{pseudo coverage}, which treats the real-time model’s prediction as a surrogate label. For batch $\mathcal{B}^t$, the pseudo coverage of real-time and pretrained CPs is
\vspace{-5px}
\begin{equation}
PC_{rt}^t = \mathbb{E}_{x \in \mathcal{B}^t} \mathds{1} \left[ 
h\left( f\left( x; \theta^t \right) \right) \in C^t_{rt}(x)
\right],
\vspace{-5px}
\end{equation}
\begin{equation}
PC_{pre}^t =\mathbb{E}_{x \in \mathcal{B}^t} \mathds{1} \left[ 
h\left( f\left( x; \theta^t \right) \right) \in C^t_{pre}(x)
\right],
\end{equation}
where $h(f(x;\theta^t))$ is the predicted label. These values act as online feedback, revealing whether each CP under- or over-covers relative to the target level.

\noindent
\textbf{Online Weight Updates}.
Guided by pseudo coverage, we update the weights $w_{rt}^t$ and $w_{pre}^t$ assigned to real-time and pretrained CPs. Instead of fixed weights, we employ an exponential update rule that decreases weights when under-coverage is detected and increases them when over-coverage occurs, thereby steering coverage towards $1-\alpha$. For the real-time CP, the update is
\vspace{-5px}
\begin{equation}
w_{rt}^t = \frac{w_{rt}^{t-1}}{T^t_{rt}},\ \ \ \ \ T^{t}_{rt} = \exp \left[ (1-\alpha) - PC_{rt}^{t-1} \right] \cdot T^{t-1}_{rt},
\vspace{-5px}
\end{equation}
with a symmetric update for $w_{pre}^t$. This enables CP to self-calibrate under evolving domains using only unlabeled feedback.

\noindent
\textbf{Coverage Analysis}.
Following the theoretical framework of weighted CP \cite{barber2023nexcptheory, farinhas2023nexcrc}, our adaptive weighting scheme satisfies the coverage bound:
\vspace{-5px}
\begin{equation}
1- \alpha - \frac{w}{nw + 1}  \sum\nolimits_{i=1}^n  \text{d}_{TV} ( Z, Z^i ) \leq 
\mathbb{P}\left(Y \in \mathcal{C}(x) \right)
\vspace{-3px}
\end{equation}
\begin{equation*}
\leq  1- \alpha + \frac{1}{nw + 1}  + \frac{w}{nw + 1} \sum\nolimits_{i=1}^n 
\text{d}_{TV}( Z, Z^i),
\end{equation*}
where $d_{TV}$ is total variation distance, $Z=((x_i,y_i))_{i=1}^{n+1}$, $Z^i$ swaps $(x_i,y_i)$ with $(x_{n+1},y_{n+1})$, and we can substitute $w$ with $w_{rt}^t$ or $w_{pre}^t$. In essence, pseudo coverage serves as an online correction signal, keeping CP coverage near the target level despite domain shift.


\subsection{Model Update}
\label{sec:modelupdate}
After each test batch is processed, the real-time model parameters are updated before the next batch arrives. 
The training objective combines cross-entropy losses from both buffers:
\vspace{-5px}
\begin{equation}
    \mathcal{L}_H ( \theta ) = \mathbb{E}_{(x, y) \in \text{Buf}_H}
    - \log ( \text{softmax} \left( f\left(x;\theta \right) \right)_y ),
\vspace{-5px}
\end{equation}
\begin{equation}
    \mathcal{L}_M ( \theta ) = \mathbb{E}_{(x, \hat{y}) \in \text{Buf}_M}
    - \log ( \text{softmax} \left( f\left(x;\theta \right) \right)_{\hat{y}}).
\end{equation}
$\text{softmax}(\cdot)_y$ denotes the softmax probability assigned to label $y$. 

%

\begin{table*}[t]
  \centering
  \caption{Real-time and post-adaptation accuracy on PACS, VLCS, and Tiny-ImageNet-C. Acc denotes the overall sample-level accuracy across all target domains or corruptions, rather than the unweighted arithmetic mean of the individual results.}
  \vspace{-10px}
  \label{tab:combined_rtpa}

  \resizebox{\textwidth}{!}{%
\begin{tabular}{ccccccccccc|ccccccccc}
\hline
\multicolumn{2}{c}{{\color[HTML]{000000} }} &
  \multicolumn{9}{c|}{\cellcolor[HTML]{FFFFFF}{\color[HTML]{000000} \textbf{PACS Dataset}}} &
  \multicolumn{9}{c}{\cellcolor[HTML]{FFFFFF}\textbf{VLCS Dataset}} \\
\multicolumn{2}{c}{{\color[HTML]{000000} }} &
  \multicolumn{4}{c}{\cellcolor[HTML]{FFE3C1}\textbf{Real-Time Accuracy}} &
  \multicolumn{5}{c|}{\cellcolor[HTML]{FFB65D}\textbf{Post-Adaptation Accuracy}} &
  \multicolumn{4}{c}{\cellcolor[HTML]{FFE3C1}\textbf{Real-Time Accuracy}} &
  \multicolumn{5}{c}{\cellcolor[HTML]{FFB65D}\textbf{Post-Adaptation Accuracy}} \\
\multicolumn{2}{c}{\multirow{-3}{*}{{\color[HTML]{000000} \textbf{Methods}}}} &
  \cellcolor[HTML]{FFE3C1}{\color[HTML]{000000} \textbf{A $\rightarrow$}} &
  \cellcolor[HTML]{FFE3C1}{\color[HTML]{000000} \textbf{$\rightarrow$ C $\rightarrow$}} &
  \cellcolor[HTML]{FFE3C1}{\color[HTML]{000000} \textbf{$\rightarrow$ S}} &
  \cellcolor[HTML]{FFE3C1}{\color[HTML]{000000} \textbf{Acc}} &
  \cellcolor[HTML]{FFB65D}{\color[HTML]{000000} \textbf{P}} &
  \cellcolor[HTML]{FFB65D}{\color[HTML]{000000} \textbf{A}} &
  \cellcolor[HTML]{FFB65D}{\color[HTML]{000000} \textbf{C}} &
  \cellcolor[HTML]{FFB65D}{\color[HTML]{000000} \textbf{S}} &
  \cellcolor[HTML]{FFB65D}{\color[HTML]{000000} \textbf{Acc}} &
  \cellcolor[HTML]{FFE3C1}\textbf{L $\rightarrow$} &
  \cellcolor[HTML]{FFE3C1}\textbf{$\rightarrow$ S $\rightarrow$} &
  \cellcolor[HTML]{FFE3C1}\textbf{$\rightarrow$ V} &
  \cellcolor[HTML]{FFE3C1}\textbf{Acc} &
  \cellcolor[HTML]{FFB65D}\textbf{C} &
  \cellcolor[HTML]{FFB65D}\textbf{L} &
  \cellcolor[HTML]{FFB65D}\textbf{S} &
  \cellcolor[HTML]{FFB65D}\textbf{V} &
  \cellcolor[HTML]{FFB65D}\textbf{Acc} \\ \hline
{\color[HTML]{000000} } &
  {\color[HTML]{000000} Tent \cite{wang2020tent}} &
  {\color[HTML]{000000} 67.53} &
  {\color[HTML]{000000} 69.16} &
  {\color[HTML]{000000} 66.94} &
  {\color[HTML]{000000} 67.71} &
  {\color[HTML]{000000} 93.53} &
  {\color[HTML]{000000} 68.46} &
  {\color[HTML]{000000} 71.25} &
  {\color[HTML]{000000} 73.12} &
  {\color[HTML]{000000} 75.14} &
  46.26 &
  41.22 &
  55.72 &
  47.91 &
  90.11 &
  52.80 &
  45.70 &
  57.08 &
  56.90 \\
{\color[HTML]{000000} } &
  {\color[HTML]{000000} CoTTA \cite{wang2022cotta}} &
  {\color[HTML]{000000} 65.72} &
  {\color[HTML]{000000} 65.91} &
  {\color[HTML]{000000} 65.28} &
  {\color[HTML]{000000} 65.57} &
  {\color[HTML]{000000} 83.71} &
  {\color[HTML]{000000} 60.60} &
  {\color[HTML]{000000} 63.48} &
  {\color[HTML]{000000} 71.24} &
  {\color[HTML]{000000} 69.32} &
  45.84 &
  39.49 &
  54.92 &
  46.89 &
  47.32 &
  52.88 &
  39.85 &
  50.24 &
  47.33 \\
{\color[HTML]{000000} } &
  {\color[HTML]{000000} SAR \cite{niu2023sar}} &
  {\color[HTML]{000000} 65.53} &
  {\color[HTML]{000000} 66.13} &
  {\color[HTML]{000000} 63.43} &
  {\color[HTML]{000000} 64.71} &
  {\color[HTML]{000000} 94.19} &
  {\color[HTML]{000000} 65.63} &
  {\color[HTML]{000000} 67.02} &
  {\color[HTML]{000000} 65.11} &
  {\color[HTML]{000000} 70.53} &
  42.69 &
  37.51 &
  50.74 &
  43.80 &
  90.47 &
  44.90 &
  38.76 &
  50.71 &
  50.86 \\
\multirow{-4}{*}{{\color[HTML]{000000} \textbf{TTA}}} &
  {\color[HTML]{000000} EATA \cite{niu2022eata}} &
  {\color[HTML]{000000} 67.43} &
  {\color[HTML]{000000} 68.26} &
  {\color[HTML]{000000} 67.24} &
  {\color[HTML]{000000} 67.57} &
  {\color[HTML]{000000} 94.85} &
  {\color[HTML]{000000} 66.31} &
  {\color[HTML]{000000} 66.72} &
  {\color[HTML]{000000} 70.60} &
  {\color[HTML]{000000} 72.86} &
  42.95 &
  37.26 &
  51.36 &
  44.01 &
  91.24 &
  43.93 &
  37.72 &
  51.74 &
  50.73 \\ \hline
{\color[HTML]{000000} } &
  {\color[HTML]{000000} SimATTA \cite{gui2024simatta}} &
  {\color[HTML]{000000} 72.56} &
  {\color[HTML]{000000} 49.36} &
  {\color[HTML]{000000} 72.49} &
  {\color[HTML]{000000} 65.99} &
  {\color[HTML]{000000} 98.68} &
  {\color[HTML]{000000} 80.91} &
  {\color[HTML]{000000} 63.99} &
  {\color[HTML]{000000} 81.37} &
  {\color[HTML]{000000} 77.85} &
  \textbf{65.38} &
  59.78 &
  63.71 &
  62.80 &
  99.86 &
  68.70 &
  65.42 &
  69.22 &
  71.97 \\
{\color[HTML]{000000} } &
  {\color[HTML]{000000} CEMA \cite{chen2024cema}} &
  {\color[HTML]{000000} 51.56} &
  {\color[HTML]{000000} 66.98} &
  {\color[HTML]{000000} 67.01} &
  {\color[HTML]{000000} 63.20} &
  {\color[HTML]{000000} 90.30} &
  {\color[HTML]{000000} 75.10} &
  {\color[HTML]{000000} 79.52} &
  {\color[HTML]{000000} 69.43} &
  {\color[HTML]{000000} 76.45} &
  55.18 &
  43.39 &
  49.35 &
  48.91 &
  78.60 &
  68.83 &
  59.63 &
  58.12 &
  63.93 \\
{\color[HTML]{000000} } &
  {\color[HTML]{000000} EATTA \cite{wang2025eatta}} &
  {\color[HTML]{000000} 70.07} &
  {\color[HTML]{000000} 69.16} &
  {\color[HTML]{000000} 65.60} &
  {\color[HTML]{000000} 67.89} &
  {\color[HTML]{000000} 98.62} &
  {\color[HTML]{000000} 72.07} &
  {\color[HTML]{000000} 71.93} &
  {\color[HTML]{000000} 68.49} &
  {\color[HTML]{000000} 75.07} &
  44.56 &
  39.48 &
  53.14 &
  45.88 &
  97.03 &
  48.10 &
  41.44 &
  53.50 &
  54.22 \\
{\color[HTML]{000000} } &
  \cellcolor[HTML]{ECF4FF}{\color[HTML]{000000} \textbf{CPATTA(Ours, $\alpha=0.1$)} } &
  \cellcolor[HTML]{ECF4FF}{\color[HTML]{000000} 71.39} &
  \cellcolor[HTML]{ECF4FF}{\color[HTML]{000000} 68.05} &
  \cellcolor[HTML]{ECF4FF}{\color[HTML]{000000} 76.18} &
  \cellcolor[HTML]{ECF4FF}{\color[HTML]{000000} 72.71} &
  \cellcolor[HTML]{ECF4FF}{\color[HTML]{000000} 98.68} &
  \cellcolor[HTML]{ECF4FF}{\color[HTML]{000000} 83.35} &
  \cellcolor[HTML]{ECF4FF}{\color[HTML]{000000} 79.27} &
  \cellcolor[HTML]{ECF4FF}{\color[HTML]{000000} 84.09} &
  \cellcolor[HTML]{ECF4FF}{\color[HTML]{000000} 85.25} &
  \cellcolor[HTML]{ECF4FF}57.99 &
  \cellcolor[HTML]{ECF4FF}\textbf{64.87} &
  \cellcolor[HTML]{ECF4FF}70.50 &
  \cellcolor[HTML]{ECF4FF}64.95 &
  \cellcolor[HTML]{ECF4FF}\textbf{100.00} &
  \cellcolor[HTML]{ECF4FF}69.46 &
  \cellcolor[HTML]{ECF4FF}72.15 &
  \cellcolor[HTML]{ECF4FF}77.34 &
  \cellcolor[HTML]{ECF4FF}76.79 \\
{\color[HTML]{000000} } &
  \cellcolor[HTML]{DAE8FC}{\color[HTML]{000000} \textbf{CPATTA(Ours, $\alpha=0.2$) }} &
  \cellcolor[HTML]{DAE8FC}{\color[HTML]{000000} \textbf{73.10}} &
  \cellcolor[HTML]{DAE8FC}{\color[HTML]{000000} 70.01} &
  \cellcolor[HTML]{DAE8FC}{\color[HTML]{000000} \textbf{79.51}} &
  \cellcolor[HTML]{DAE8FC}{\color[HTML]{000000} \textbf{75.26}} &
  \cellcolor[HTML]{DAE8FC}{\color[HTML]{000000} \textbf{99.10}} &
  \cellcolor[HTML]{DAE8FC}{\color[HTML]{000000} \textbf{85.55}} &
  \cellcolor[HTML]{DAE8FC}{\color[HTML]{000000} 80.29} &
  \cellcolor[HTML]{DAE8FC}{\color[HTML]{000000} \textbf{86.94}} &
  \cellcolor[HTML]{DAE8FC}{\color[HTML]{000000} \textbf{87.13}} &
  \cellcolor[HTML]{DAE8FC} 58.11 &
  \cellcolor[HTML]{DAE8FC}61.88 &
  \cellcolor[HTML]{DAE8FC}\textbf{73.34} &
  \cellcolor[HTML]{DAE8FC}\textbf{64.96} &
  \cellcolor[HTML]{DAE8FC}99.58 &
  \cellcolor[HTML]{DAE8FC}\textbf{70.17} &
  \cellcolor[HTML]{DAE8FC}\textbf{72.58} &
  \cellcolor[HTML]{DAE8FC}\textbf{79.47} &
  \cellcolor[HTML]{DAE8FC}\textbf{77.72} \\
\multirow{-6}{*}{{\color[HTML]{000000} \textbf{ATTA}}} &
  \cellcolor[HTML]{C4DDFF}{\color[HTML]{000000} \textbf{CPATTA(Ours, $\alpha=0.3$)} } &
  \cellcolor[HTML]{C4DDFF}{\color[HTML]{000000} 72.51} &
  \cellcolor[HTML]{C4DDFF}{\color[HTML]{000000} \textbf{73.76}} &
  \cellcolor[HTML]{C4DDFF}{\color[HTML]{000000} 74.60} &
  \cellcolor[HTML]{C4DDFF}{\color[HTML]{000000} 73.85} &
  \cellcolor[HTML]{C4DDFF}{\color[HTML]{000000} 97.96} &
  \cellcolor[HTML]{C4DDFF}{\color[HTML]{000000} 83.40} &
  \cellcolor[HTML]{C4DDFF}{\color[HTML]{000000} \textbf{82.89}} &
  \cellcolor[HTML]{C4DDFF}{\color[HTML]{000000} 83.33} &
  \cellcolor[HTML]{C4DDFF}{\color[HTML]{000000} 85.69} &
  \cellcolor[HTML]{C4DDFF}57.77 &
  \cellcolor[HTML]{C4DDFF}62.52 &
  \cellcolor[HTML]{C4DDFF}72.66 &
  \cellcolor[HTML]{C4DDFF}64.84 &
  \cellcolor[HTML]{C4DDFF}99.93 &
  \cellcolor[HTML]{C4DDFF}69.83 &
  \cellcolor[HTML]{C4DDFF}71.21 &
  \cellcolor[HTML]{C4DDFF}76.57 &
  \cellcolor[HTML]{C4DDFF}76.35 \\ \hline
\end{tabular}
  }

  \vspace{0mm}

  \resizebox{\textwidth}{!}{%
\begin{tabular}{ccccccccccccccccc}
\hline
\rowcolor[HTML]{FFFFFF} 
\cellcolor[HTML]{FFFFFF}{\color[HTML]{000000} } &
  \multicolumn{15}{c}{\cellcolor[HTML]{FFFFFF}{\color[HTML]{000000} \begin{tabular}[c]{@{}c@{}}\textbf{Real-Time Accuracy}\end{tabular}}} &
  {\color[HTML]{000000} \begin{tabular}[c]{@{}c@{}} \textbf{Post-Adapt. Acc.} \end{tabular}} \\
\cellcolor[HTML]{FFFFFF}{\color[HTML]{000000} } &
  \multicolumn{3}{c}{\cellcolor[HTML]{FFF2E2}{\color[HTML]{000000} \textbf{Noise $\rightarrow$}}} &
  \multicolumn{4}{c}{\cellcolor[HTML]{FFE3C1}{\color[HTML]{000000} \textbf{$\rightarrow$Blur$\rightarrow$}}} &
  \multicolumn{3}{c}{\cellcolor[HTML]{FFCE93}{\color[HTML]{000000} \textbf{$\rightarrow$Weather$\rightarrow$}}} &
  \multicolumn{4}{c}{\cellcolor[HTML]{FFB65D}{\color[HTML]{000000} \textbf{$\rightarrow$Digital}}} &
  \cellcolor[HTML]{FFFFFF}{\color[HTML]{000000} } &
  \cellcolor[HTML]{FFFFFF}{\color[HTML]{000000} } \\
\multirow{-3}{*}{\cellcolor[HTML]{FFFFFF}{\color[HTML]{000000} \textbf{Methods}}} &
  \cellcolor[HTML]{FFF2E2}{\color[HTML]{000000} \textbf{Gauss.}} &
  \cellcolor[HTML]{FFF2E2}{\color[HTML]{000000} \textbf{Shot}} &
  \cellcolor[HTML]{FFF2E2}{\color[HTML]{000000} \textbf{Impul.}} &
  \cellcolor[HTML]{FFE3C1}{\color[HTML]{000000} \textbf{Defoc.}} &
  \cellcolor[HTML]{FFE3C1}{\color[HTML]{000000} \textbf{Glass}} &
  \cellcolor[HTML]{FFE3C1}{\color[HTML]{000000} \textbf{Motion}} &
  \cellcolor[HTML]{FFE3C1}{\color[HTML]{000000} \textbf{Zoom}} &
  \cellcolor[HTML]{FFCE93}{\color[HTML]{000000} \textbf{Snow}} &
  \cellcolor[HTML]{FFCE93}{\color[HTML]{000000} \textbf{Frost}} &
  \cellcolor[HTML]{FFCE93}{\color[HTML]{000000} \textbf{Fog}} &
  \cellcolor[HTML]{FFB65D}{\color[HTML]{000000} \textbf{Contr.}} &
  \cellcolor[HTML]{FFB65D}{\color[HTML]{000000} \textbf{Elast.}} &
  \cellcolor[HTML]{FFB65D}{\color[HTML]{000000} \textbf{Pixel}} &
  \cellcolor[HTML]{FFB65D}{\color[HTML]{000000} \textbf{JPEG}} &
  \multirow{-2}{*}{\cellcolor[HTML]{FFFFFF}{\color[HTML]{000000} \textbf{Avg}}} &
  \multirow{-2}{*}{\cellcolor[HTML]{FFFFFF}{\color[HTML]{000000} \textbf{Avg}}} \\ \hline
\rowcolor[HTML]{FFFFFF} 
{\color[HTML]{000000} Tent \cite{wang2020tent}} &
  {\color[HTML]{000000} 6.57} &
  {\color[HTML]{000000} 7.69} &
  {\color[HTML]{000000} 6.81} &
  {\color[HTML]{000000} 11.49} &
  {\color[HTML]{000000} 6.51} &
  {\color[HTML]{000000} 17.41} &
  {\color[HTML]{000000} 15.92} &
  {\color[HTML]{000000} 17.83} &
  {\color[HTML]{000000} 17.52} &
  {\color[HTML]{000000} 16.28} &
  {\color[HTML]{000000} \textbf{4.64}} &
  {\color[HTML]{000000} 14.61} &
  {\color[HTML]{000000} 17.71} &
  {\color[HTML]{000000} 16.62} &
  {\color[HTML]{000000} 12.69} &
  \cellcolor[HTML]{FFFFFF}{\color[HTML]{000000} 13.86} \\
\rowcolor[HTML]{FFFFFF} 
{\color[HTML]{000000} CoTTA \cite{wang2022cotta}} &
  {\color[HTML]{000000} 3.38} &
  {\color[HTML]{000000} 2.96} &
  {\color[HTML]{000000} 2.56} &
  {\color[HTML]{000000} 3.28} &
  {\color[HTML]{000000} 2.25} &
  {\color[HTML]{000000} 4.51} &
  {\color[HTML]{000000} 4.02} &
  {\color[HTML]{000000} 4.68} &
  {\color[HTML]{000000} 4.76} &
  {\color[HTML]{000000} 3.16} &
  {\color[HTML]{000000} 1.66} &
  {\color[HTML]{000000} 3.16} &
  {\color[HTML]{000000} 3.77} &
  {\color[HTML]{000000} 4.04} &
  {\color[HTML]{000000} 3.44} &
  \cellcolor[HTML]{FFFFFF}{\color[HTML]{000000} 9.10} \\
\rowcolor[HTML]{FFFFFF} 
{\color[HTML]{000000} SAR \cite{niu2023sar}} &
  {\color[HTML]{000000} 6.35} &
  {\color[HTML]{000000} 6.91} &
  {\color[HTML]{000000} 6.10} &
  {\color[HTML]{000000} 9.81} &
  {\color[HTML]{000000} 5.40} &
  {\color[HTML]{000000} 15.13} &
  {\color[HTML]{000000} 13.61} &
  {\color[HTML]{000000} 16.31} &
  {\color[HTML]{000000} 15.44} &
  {\color[HTML]{000000} 14.47} &
  {\color[HTML]{000000} 3.18} &
  {\color[HTML]{000000} 12.87} &
  {\color[HTML]{000000} 14.79} &
  {\color[HTML]{000000} 15.30} &
  {\color[HTML]{000000} 11.12} &
  \cellcolor[HTML]{FFFFFF}{\color[HTML]{000000} 12.61} \\
\rowcolor[HTML]{FFFFFF} 
{\color[HTML]{000000} EATA \cite{niu2022eata}} &
  {\color[HTML]{000000} 6.30} &
  {\color[HTML]{000000} 6.85} &
  {\color[HTML]{000000} 6.00} &
  {\color[HTML]{000000} 9.71} &
  {\color[HTML]{000000} 5.31} &
  {\color[HTML]{000000} 14.94} &
  {\color[HTML]{000000} 13.67} &
  {\color[HTML]{000000} 16.38} &
  {\color[HTML]{000000} 15.39} &
  {\color[HTML]{000000} 14.14} &
  {\color[HTML]{000000} 3.11} &
  {\color[HTML]{000000} 12.76} &
  {\color[HTML]{000000} 14.72} &
  {\color[HTML]{000000} 14.88} &
  {\color[HTML]{000000} 11.01} &
  \cellcolor[HTML]{FFFFFF}{\color[HTML]{000000} 12.47} \\ \hline
\rowcolor[HTML]{FFFFFF} 
{\color[HTML]{000000} SimATTA \cite{gui2024simatta}} &
  {\color[HTML]{000000} \textbf{15.77}} &
  {\color[HTML]{000000} 28.31} &
  {\color[HTML]{000000} 15.17} &
  {\color[HTML]{000000} 22.61} &
  {\color[HTML]{000000} 17.80} &
  {\color[HTML]{000000} 44.10} &
  {\color[HTML]{000000} 44.32} &
  {\color[HTML]{000000} 38.67} &
  {\color[HTML]{000000} 49.32} &
  {\color[HTML]{000000} 30.41} &
  {\color[HTML]{000000} 1.88} &
  {\color[HTML]{000000} 41.1} &
  {\color[HTML]{000000} 55.30} &
  {\color[HTML]{000000} 61.38} &
  {\color[HTML]{000000} 33.30} &
  \cellcolor[HTML]{FFFFFF}{\color[HTML]{000000} 41.81} \\
\rowcolor[HTML]{FFFFFF} 
{\color[HTML]{000000} CEMA \cite{chen2024cema}} &
  {\color[HTML]{000000} 10.65} &
  {\color[HTML]{000000} \textbf{28.96}} &
  {\color[HTML]{000000} \textbf{22.68}} &
  {\color[HTML]{000000} 23.79} &
  {\color[HTML]{000000} 16.90} &
  {\color[HTML]{000000} 40.45} &
  {\color[HTML]{000000} 40.21} &
  {\color[HTML]{000000} 34.85} &
  {\color[HTML]{000000} 43.85} &
  {\color[HTML]{000000} 36.07} &
  {\color[HTML]{000000} 2.79} &
  {\color[HTML]{000000} 37.18} &
  {\color[HTML]{000000} 55.71} &
  {\color[HTML]{000000} 55.21} &
  {\color[HTML]{000000} 32.09} &
  \cellcolor[HTML]{FFFFFF}{\color[HTML]{000000} 41.57} \\
\rowcolor[HTML]{FFFFFF} 
{\color[HTML]{000000} EATTA \cite{wang2025eatta}} &
  {\color[HTML]{000000} 7.81} &
  {\color[HTML]{000000} 8.31} &
  {\color[HTML]{000000} 6.89} &
  {\color[HTML]{000000} 12.18} &
  {\color[HTML]{000000} 6.41} &
  {\color[HTML]{000000} 18.61} &
  {\color[HTML]{000000} 16.88} &
  {\color[HTML]{000000} 20.31} &
  {\color[HTML]{000000} 19.11} &
  {\color[HTML]{000000} 16.99} &
  {\color[HTML]{000000} 3.90} &
  {\color[HTML]{000000} 16.00} &
  {\color[HTML]{000000} 18.64} &
  {\color[HTML]{000000} 19.12} &
  {\color[HTML]{000000} 13.65} &
  \cellcolor[HTML]{FFFFFF}{\color[HTML]{000000} 15.57} \\
\rowcolor[HTML]{ECF4FF} 
{\color[HTML]{000000} \textbf{CPATTA(Ours, $\alpha=0.3$)} }&
  {\color[HTML]{000000} 13.88} &
  {\color[HTML]{000000} 27.18} &
  {\color[HTML]{000000} 19.99} &
  {\color[HTML]{000000} \textbf{24.36}} &
  {\color[HTML]{000000} \textbf{17.82}} &
  {\color[HTML]{000000} \textbf{44.39}} &
  {\color[HTML]{000000} \textbf{44.63}} &
  {\color[HTML]{000000} 39.69} &
  {\color[HTML]{000000} 49.24} &
  {\color[HTML]{000000} \textbf{37.30}} &
  {\color[HTML]{000000} 2.57} &
  {\color[HTML]{000000} 42.50} &
  {\color[HTML]{000000} \textbf{61.05}} &
  {\color[HTML]{000000} \textbf{62.15}} &
  {\color[HTML]{000000} \textbf{34.77}} &
  {\color[HTML]{000000} \textbf{44.59}} \\
\rowcolor[HTML]{DAE8FC} 
{\color[HTML]{000000} \textbf{CPATTA(Ours, $\alpha=0.4$)}} &
  {\color[HTML]{000000} 13.88} &
  {\color[HTML]{000000} 27.18} &
  {\color[HTML]{000000} 19.99} &
  {\color[HTML]{000000} \textbf{24.36}} &
  {\color[HTML]{000000} \textbf{17.82}} &
  {\color[HTML]{000000} \textbf{44.39}} &
  {\color[HTML]{000000} \textbf{44.63}} &
  {\color[HTML]{000000} \textbf{40.03}} &
  {\color[HTML]{000000} \textbf{50.18}} &
  {\color[HTML]{000000} 36.93} &
  {\color[HTML]{000000} 2.73} &
  {\color[HTML]{000000} 42.01} &
  {\color[HTML]{000000} 59.97} &
  {\color[HTML]{000000} 62.00} &
  {\color[HTML]{000000} 34.72} &
  {\color[HTML]{000000} 43.94} \\
\rowcolor[HTML]{C4DDFF} 
{\color[HTML]{000000} \textbf{ CPATTA(Ours, $\alpha=0.5$)} } &
  {\color[HTML]{000000} 13.88} &
  {\color[HTML]{000000} 27.18} &
  {\color[HTML]{000000} 19.99} &
  {\color[HTML]{000000} \textbf{24.36}} &
  {\color[HTML]{000000} \textbf{17.82}} &
  {\color[HTML]{000000} \textbf{44.39}} &
  {\color[HTML]{000000} \textbf{44.63}} &
  {\color[HTML]{000000} 39.69} &
  {\color[HTML]{000000} 49.33} &
  {\color[HTML]{000000} 36.36} &
  {\color[HTML]{000000} 1.70} &
  {\color[HTML]{000000} \textbf{43.06}} &
  {\color[HTML]{000000} 60.73} &
  {\color[HTML]{000000} 62.06} &
  {\color[HTML]{000000} 34.66} &
  {\color[HTML]{000000} 43.85} \\ \hline
\end{tabular}
  }
\end{table*}

To balance reliability and efficiency, we adopt a two-stage update with buffer-specific learning rates. First, parameters are updated using human-labeled data with rate $\eta_H$
\vspace{-5px}
\begin{equation}
\theta^{t + \frac{1}{2}} = \theta^t - \eta_H \nabla_{\theta} \mathcal{L}_H \left( \theta^t \right),
\vspace{-5px}
\end{equation}
and then further updated with model-labeled data using rate $\eta_M$:
\vspace{-5px}
\begin{equation}
\theta^{t + 1} = \theta^{t + \frac{1}{2}} - \eta_M \nabla_{\theta} \mathcal{L}_M \left( \theta^{t+\frac{1}{2}} \right).
\vspace{-5px}
\end{equation}
This staged update scheme reflects the design principle of trust but expand. Reliable human supervision anchors the model update, while additional pseudo-labeled data broaden adaptation at low cost. The method ensures that human annotations dominate parameter correction, whereas model labels serve as auxiliary signals that accelerate adaptation without overwhelming reliability. 

\vspace{-10px}
\subsection{Discussion: NexCP and Source-Access Fairness}

\noindent\textbf{Nonexchangeable CP.}
Practical nonexchangeable CP schemes often use fixed or heuristic weights \cite{barber2023nexcptheory,farinhas2023nexcrc}, while QTC \cite{yilmaz2022qtc} relies on model predictions remaining reliable when adjusting conformal thresholds. Neither approach directly accounts for the simultaneous evolution of target distributions and model parameters in ATTA. CPATTA instead uses pseudo coverage as online feedback to adapt conformal weights across batches, providing feedback-based coverage adjustment under continual shift without claiming classical distribution-free validity.

\noindent\textbf{Source-access fairness.}
CPATTA explicitly considers source-calibrated rather than strict source-free ATTA, consistent with prior TTA methods that retain lightweight source-derived statistics or proxies \cite{kang2023fairness1,adachi2023fairness2}. The retained samples are used only for conformal calibration, never for gradient updates, source retraining, or replay. As a controlled comparison, Table~\ref{tab:replay} gives competing ATTA methods the identical samples as labeled replay data, a more direct use of source supervision; CPATTA nevertheless remains superior. Thus, source access alone does not explain the gains, supporting the benefit of calibration-aware supervision allocation.

\section{Experiments}
\textbf{Experiment Details}. 
We evaluate on three domain-shift benchmarks: PACS, VLCS, and Tiny-ImageNet-C. PACS \cite{li2017pacs} contains 7 classes across 4 domains: Photo, Art Painting, Cartoon, and Sketch. VLCS \cite{fang2013vlcs} combines 4 datasets with 5 shared classes, producing natural cross-domain shifts. Tiny-ImageNet-C \cite{hendrycks2018tinyimagenetc} applies 15 corruption types at 5 severity levels to the 200-class Tiny-ImageNet; we use severity 5. We report real-time accuracy, post-adaptation accuracy, and coverage gap. Calibration sets contain 50 source-domain images per class for PACS and VLCS (350 and 250 images, respectively) and 3 per class for Tiny-ImageNet-C (600 images). \textbf{For human annotation, we select the 3 most uncertain samples per batch on PACS/VLCS and 2 on Tiny-ImageNet-C, with budgets capped at 300 and 3,000, respectively, for all methods.} A ResNet-18 backbone \cite{he2016resnet} was pretrained on domains \textit{P}, \textit{C}, and \textit{brightness} for PACS, VLCS, and Tiny-ImageNet-C, respectively.

\noindent
\textbf{Compared Methods}.
We compare CPATTA with four TTA methods—Tent \cite{wang2020tent}, CoTTA \cite{wang2022cotta}, SAR \cite{niu2023sar}, and EATA \cite{niu2022eata}—and three ATTA methods—CEMA \cite{chen2024cema}, SimATTA \cite{gui2024simatta}, and EATTA \cite{wang2025eatta}. Following \cite{gui2024simatta}, TTA baselines receive randomly sampled labels under the same annotation budget as ATTA methods. For CEMA, we retain selective querying but replace foundation-model annotations with human labels.

\noindent
\textbf{Real-Time and Post-Adaptation Accuracy}. 
Table \ref{tab:combined_rtpa} reports real-time and post-adaptation accuracies across different miscoverage levels: $\alpha \in \{0.1, 0.2, 0.3\}$ for PACS and VLCS, and $\alpha \in \{0.3, 0.4, 0.5\}$ for Tiny-ImageNet-C, which yields stronger performance. CPATTA consistently outperforms existing methods. At $\alpha=0.2$, it achieves 75.26\%/87.13\% on PACS and 64.96\%/77.72\% on VLCS, substantially surpassing SimATTA, CEMA, and EATTA. On Tiny-ImageNet-C, CPATTA obtains the best real-time/post-adaptation accuracies of 34.77\%/44.59\%, demonstrating strong robustness across diverse distribution shifts.

\begin{table}[t]
\vspace{-15px}
\caption{Data selection efficiencies with $\alpha \in \{0.1, 0.2, 0.3\}$ for PACS/VLCS and $\alpha \in \{0.3, 0.4, 0.5\}$ for Tiny-ImageNet-C}
\vspace{-10px}
\centering
\resizebox{\columnwidth}{!}{%

\begin{tabular}{c|cc|cc|cc}
\hline
{\color[HTML]{000000} } &
  \multicolumn{2}{c|}{{\color[HTML]{000000} \textbf{PACS}}} &
  \multicolumn{2}{c|}{{\color[HTML]{000000} \textbf{VLCS}}} &
  \multicolumn{2}{c}{{\color[HTML]{000000} \textbf{Tiny-ImageNet-C}}} \\
\multirow{-2}{*}{{\color[HTML]{000000} Method}} &
  {\color[HTML]{000000} \textbf{$\text{Eff}_H$}} &
  {\color[HTML]{000000} \textbf{$\text{Eff}_M$}} &
  {\color[HTML]{000000} \textbf{$\text{Eff}_H$}} &
  {\color[HTML]{000000} \textbf{$\text{Eff}_M$}} &
  {\color[HTML]{000000} \textbf{$\text{Eff}_H$}} &
  {\color[HTML]{000000} \textbf{$\text{Eff}_M$}} \\ \hline
{\color[HTML]{000000} SimATTA \cite{gui2024simatta}} &
  {\color[HTML]{000000} 47.60} &
  {\color[HTML]{000000} 67.60} &
  {\color[HTML]{000000} 57.04} &
  {\color[HTML]{000000} 64.78} &
  {\color[HTML]{000000} 79.25} &
  {\color[HTML]{000000} \textbf{50.00}} \\
{\color[HTML]{000000} CEMA \cite{chen2024cema}} &
  {\color[HTML]{000000} 24.91} &
  {\color[HTML]{000000} N/A} &
  {\color[HTML]{000000} 44.38} &
  {\color[HTML]{000000} N/A} &
  {\color[HTML]{000000} 75.82} &
  {\color[HTML]{000000} N/A} \\
{\color[HTML]{000000} EATTA \cite{wang2025eatta}} &
  {\color[HTML]{000000} 53.71} &
  {\color[HTML]{000000} 90.71} &
  {\color[HTML]{000000} 58.38} &
  {\color[HTML]{000000} 63.60} &
  {\color[HTML]{000000} 62.90} &
  {\color[HTML]{000000} 32.31} \\
\rowcolor[HTML]{ECF4FF} 
{\color[HTML]{000000} CPATTA($\alpha = 0.1 / 0.3$)} &
  {\color[HTML]{000000} 66.30} &
  {\color[HTML]{000000} 93.41} &
  {\color[HTML]{000000} 62.89} &
  {\color[HTML]{000000} 81.91} &
  {\color[HTML]{000000} 90.45} &
  {\color[HTML]{000000} 48.14} \\
\rowcolor[HTML]{DAE8FC} 
{\color[HTML]{000000} CPATTA($\alpha = 0.2 / 0.4$)} &
  {\color[HTML]{000000} 65.94} &
  {\color[HTML]{000000} 93.65} &
  {\color[HTML]{000000} 61.86} &
  {\color[HTML]{000000} \textbf{85.11}} &
  {\color[HTML]{000000} \textbf{91.00}} &
  {\color[HTML]{000000} 47.83} \\
\rowcolor[HTML]{C4DDFF} 
{\color[HTML]{000000} CPATTA($\alpha = 0.3 / 0.5$)} &
  {\color[HTML]{000000} \textbf{67.39}} &
  {\color[HTML]{000000} \textbf{94.59}} &
  {\color[HTML]{000000} \textbf{64.29}} &
  {\color[HTML]{000000} 84.04} &
  {\color[HTML]{000000} 90.08} &
  {\color[HTML]{000000} 47.93} \\ \hline
\end{tabular}

}

\vspace{-10px}

\label{tab:data_se}

\end{table}
\begin{table}[h]
\caption{Ablation under the same human annotation budget}
\vspace{-10px}
\label{tab:component-ablation}
\resizebox{\linewidth}{!}{
\begin{tabular}{ccccccc}
\hline
\cellcolor[HTML]{FFFFFF}{\color[HTML]{000000} } &
  \multicolumn{3}{c}{\cellcolor[HTML]{FFE3C1}{\color[HTML]{000000} Realtime Accuracy}} &
  \multicolumn{3}{c}{\cellcolor[HTML]{FFB65D}{\color[HTML]{000000} Post-Adaptation Accuracy}} \\
\multirow{-2}{*}{\cellcolor[HTML]{FFFFFF}{\color[HTML]{000000} \textbf{Variants}}} &
  \cellcolor[HTML]{FFE3C1}{\color[HTML]{000000} \textbf{PACS}} &
  \cellcolor[HTML]{FFE3C1}{\color[HTML]{000000} \textbf{VLCS}} &
  \cellcolor[HTML]{FFE3C1}{\color[HTML]{000000} \textbf{Tiny}} &
  \cellcolor[HTML]{FFB65D}{\color[HTML]{000000} \textbf{PACS}} &
  \cellcolor[HTML]{FFB65D}{\color[HTML]{000000} \textbf{VLCS}} &
  \cellcolor[HTML]{FFB65D}{\color[HTML]{000000} \textbf{Tiny}} \\ \hline
\\[-0.8em]
{\color[HTML]{000000} Random Selection  } &
  {\color[HTML]{000000}68.16} &
  {\color[HTML]{000000}55.58} &
  {\color[HTML]{000000}31.98} &
  {\color[HTML]{000000}78.37} &
  {\color[HTML]{000000}70.77} &
  {\color[HTML]{000000}40.96} \\ 

{\color[HTML]{000000} w/o adaptive weighting (Gometric Decaying)} &
  {\color[HTML]{000000}74.31} &
  {\color[HTML]{000000}63.29} &
  {\color[HTML]{000000}29.22} &
  {\color[HTML]{000000}83.37} &
  {\color[HTML]{000000}75.14} &
  {\color[HTML]{000000}37.14} \\ 

{\color[HTML]{000000} Fixed hard-set size } &
  {\color[HTML]{000000}72.02} &
  {\color[HTML]{000000}58.34} &
  {\color[HTML]{000000}29.38} &
  {\color[HTML]{000000}83.09} &
  {\color[HTML]{000000}68.03} &
  {\color[HTML]{000000}37.58} \\ 

  {\color[HTML]{000000} w/o DSS-triggered budget adjustment } &
  {\color[HTML]{000000}68.50} &
  {\color[HTML]{000000}64.64} &
  {\color[HTML]{000000}30.69} &
  {\color[HTML]{000000}84.83} &
  {\color[HTML]{000000}74.91} &
  {\color[HTML]{000000}38.45} \\ 

  {\color[HTML]{000000} Joint human--pseudo update } &
  {\color[HTML]{000000}69.53} &
  {\color[HTML]{000000}57.01} &
  {\color[HTML]{000000}28.32} &
  {\color[HTML]{000000}84.70} &
  {\color[HTML]{000000}72.78} &
  {\color[HTML]{000000}36.38} \\ 
\rowcolor[HTML]{C4DDFF} 
  {\color[HTML]{000000} Full CPATTA } &
  {\color[HTML]{000000}\textbf{75.26}} &
  {\color[HTML]{000000}\textbf{64.96}} &
  {\color[HTML]{000000}\textbf{34.77}} &
  {\color[HTML]{000000}\textbf{87.13}} &
  {\color[HTML]{000000}\textbf{77.72}} &
  {\color[HTML]{000000}\textbf{44.59}} \\ \hline

\end{tabular}}
\vspace{-20px}
\end{table}

\noindent
\textbf{Data Selection Efficiency}. We formally define efficiency of human annotation($\text{Eff}_H$) and model annotation($\text{Eff}_M$) as follows:
\vspace{-5px}
\begin{equation}
\text{Eff}_{H} = \mathbb{E}_{ (x, y) \in \text{Buf}_H} \mathds{1} \left[ 
y \neq \hat{y}
\right],\ \ \ \ \text{(Model Predicts Wrong)}
\vspace{-5px}
\end{equation}
\begin{equation}
\text{Eff}_{M} = \mathbb{E}_{ \left( x, \hat{y} \right) \in \text{Buf}_M} \mathds{1} \left[ 
y = \hat{y}
\right].\ \ \ \ \text{(Model Predicts Right)}
\end{equation}
As shown in Table \ref{tab:data_se}, CPATTA achieves the highest $\text{Eff}_H$ across all datasets, while maintaining competitive or superior $\text{Eff}_M$. The results are stable across different $\alpha$ values, demonstrating CPATTA’s effectiveness in improving annotation efficiency in ATTA.

\begin{figure}[t]
    \centering
    \vspace{-15px}
    \begin{subfigure}[t]{\columnwidth}
        \centering
        \includegraphics[width=\linewidth]{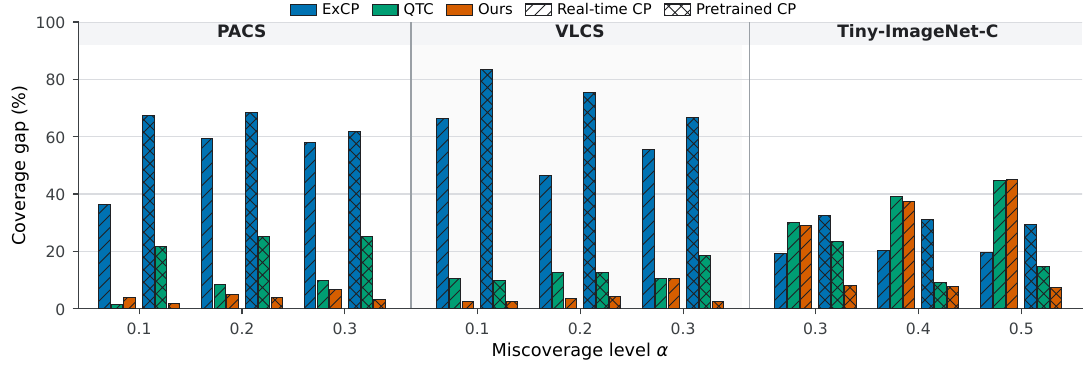}
        \label{fig:coverage-gap}
    \end{subfigure}

    \vspace{-4mm}

    \begin{subfigure}[t]{\columnwidth}
        \centering
        \includegraphics[width=\linewidth]{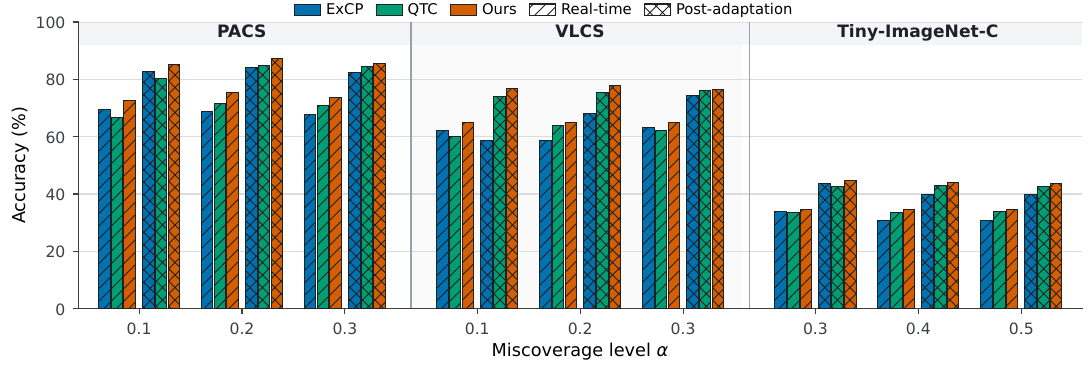}
        \label{fig:adaptation-accuracy}
    \end{subfigure}
    \vspace{-20px}
    \caption{Compare coverage gap and adaptation acc with other CPs.}
    \vspace{-15px}
    \label{fig:combined-results-vertical}
\end{figure}

\noindent
\textbf{Coverage Gap}. 
Figure~\ref{fig:combined-results-vertical} compares the coverage gap and predictive accuracy of exchangeable CP (ExCP), QTC, and our method across PACS, VLCS, and Tiny-ImageNet-C. Our method achieves the smallest or near-smallest coverage gaps in most settings, with particularly consistent improvements when using pretrained CP. On Tiny-ImageNet-C, ExCP provides smaller real-time coverage gaps in some cases, reflecting the difficulty of calibration under stronger distribution shifts.

For both real-time and post-adaptation evaluation, our method achieves the highest or comparable accuracy across datasets and miscoverage levels. Overall, the results show that our method provides a favorable balance between conformal calibration and predictive performance compared with existing conformal prediction methods.

\noindent
\textbf{Ablation Study.} We conduct ablation studies on three datasets under the same human annotation budget, as shown in Table~\ref{tab:component-ablation}. Full CPATTA achieves the highest real-time and post-adaptation accuracy in all settings, outperforming random selection by up to 9.38 and 8.76 points, respectively. Removing adaptive weighting particularly affects Tiny, while using a fixed hard-set size, disabling DSS-triggered budget adjustment, and replacing staged updates with joint updates consistently degrade performance. These results demonstrate the complementary contribution of each component.

\begin{table}[!h]
\vspace{-10px}
\centering
\caption{Replay Experiment on PACS}
\vspace{-10px}
\resizebox{\columnwidth}{!}{%

\begin{tabular}{c|cccc|c}
\hline
{\color[HTML]{000000} } &
  \multicolumn{4}{c|}{{\color[HTML]{000000} \textbf{Real-Time}}} &
  {\color[HTML]{000000} \textbf{Post-Adapt}} \\
\multirow{-2}{*}{{\color[HTML]{000000} \textbf{Method}}} &
  {\color[HTML]{000000} \textbf{A $\rightarrow$}} &
  {\color[HTML]{000000} \textbf{$\rightarrow$ C $\rightarrow$}} &
  {\color[HTML]{000000} \textbf{$\rightarrow$ S}} &
  {\color[HTML]{000000} \textbf{Acc}} &
  {\color[HTML]{000000} \textbf{Acc}} \\ \hline
{\color[HTML]{000000} SimATTA with Replay} &
  {\color[HTML]{000000} \textbf{74.32}} &
  {\color[HTML]{000000} 50.64} &
  {\color[HTML]{000000} 72.77} &
  {\color[HTML]{000000} 66.92} &
  {\color[HTML]{000000} 79.14} \\
CEMA with Replay &
  65.58 &
  68.64 &
  58.28 &
  63.00 &
  75.38 \\
EATTA with Replay &
  69.97 &
  69.07 &
  65.59 &
  67.65 &
  74.85 \\
\rowcolor[HTML]{ECF4FF} 
{\color[HTML]{000000} CPATTA($\alpha = 0.1$)} &
  {\color[HTML]{000000} 71.39} &
  {\color[HTML]{000000} 68.05} &
  {\color[HTML]{000000} 76.18} &
  {\color[HTML]{000000} 72.71} &
  {\color[HTML]{000000} 85.25} \\
\rowcolor[HTML]{DAE8FC} 
{\color[HTML]{000000} CPATTA($\alpha = 0.2$)} &
  {\color[HTML]{000000} 73.10} &
  {\color[HTML]{000000} 70.01} &
  {\color[HTML]{000000} \textbf{79.51}} &
  {\color[HTML]{000000} \textbf{75.26}} &
  {\color[HTML]{000000} \textbf{87.13}} \\
\rowcolor[HTML]{C4DDFF} 
{\color[HTML]{000000} CPATTA($\alpha = 0.3$)} &
  {\color[HTML]{000000} 72.51} &
  {\color[HTML]{000000} \textbf{73.76}} &
  {\color[HTML]{000000} 74.60} &
  {\color[HTML]{000000} 73.85} &
  {\color[HTML]{000000} 85.69} \\ \hline
\end{tabular}

}
\label{tab:replay}
\vspace{-10px}
\end{table}

\noindent \textbf{Replay Experiment}. To further show the effectiveness of CP in ATTA, we allow the compared methods access to the labeled calibration set as replay data during adaptation. Table \ref{tab:replay} reports results on PACS.  While SimATTA with replay improves accuracy on domain \textit{A}, CPATTA achieves superior performance on the other two domains and obtains the best overall real-time and post-adaptation accuracy. This confirms that leveraging the calibration set through CP is more effective than using it as replay data.

{
\setlength{\columnsep}{4pt} 
\setlength{\intextsep}{2pt} 

\begin{wraptable}{r}{0.50\columnwidth} 
\centering
\caption{Pseudo-coverage validation on PACS and VLCS.}
\label{tab:pseudo-coverage-validation}
\vspace{-10pt}

\scriptsize
\setlength{\tabcolsep}{0.8 pt}
\renewcommand{\arraystretch}{1.3}

\begin{tabular}{@{}ccccc}
\hline
\multirow{2}{*}{\textbf{Metric}} &
\multicolumn{2}{c}{\cellcolor[HTML]{FFE3C1}\textbf{Real-Time CP}} &
\multicolumn{2}{c}{\cellcolor[HTML]{FFB65D}\textbf{Pretrained CP}} \\
&
\cellcolor[HTML]{FFE3C1}\textbf{PACS}&
\cellcolor[HTML]{FFE3C1}\textbf{VLCS}&
\cellcolor[HTML]{FFB65D}\textbf{PACS}&
\cellcolor[HTML]{FFB65D}\textbf{VLCS}\\
\hline
Corr.    & 0.9978 & 0.9997 & 0.9988 & 0.9853 \\
\hline
\end{tabular}
\end{wraptable}

\noindent\textbf{Reliability of Pseudo Coverage.}
Under domain shift, weighted CP requires exact importance weights for exact coverage \cite{barber2023nexcptheory,farinhas2023nexcrc}, which are unavailable in online ATTA with an unlabeled target stream and limited source calibration data. CPATTA therefore uses pseudo coverage as a label-free signal for adapting conformal weights, rather than claiming exact coverage. We validate this proxy offline by running CPATTA without target labels and revealing benchmark labels only after logging. Table~\ref{tab:pseudo-coverage-validation} shows high pseudo--true coverage correlations, supporting its use as an online control signal rather than an oracle estimate.
\par
}

\vspace{-5px}
\section{Conclusion}
\vspace{-5px}
CP provides a principled uncertainty measure for selecting data for human and model annotation in ATTA. To address its coverage gap under domain shift, we propose adaptive weighting that improves robustness and calibration across domains. Our method uses annotated data more efficiently, conserving scarce supervision while enabling reliable pseudo-labels and stronger adaptation.
Experiments show CPATTA achieves roughly 5\% higher accuracy than SOTA ATTA methods. Future work will explore improved update strategies and statistical modeling of annotation errors in deployments.

\clearpage

\noindent \textbf{Funding Acknowledgments:} This work was supported by National Key R\&D Program of China 2022YFC3400400, 2023YFC3503400; NSFC-FDCT Grants 62361166662; The Innovative Research Group Project of Hunan Province 2024JJ1002; Hunan Science and Technology Innovation Plan 2025ZYJ003; Key R\&D Program of Hunan Province 2023GK2004, 2023SK2059, 2023SK2060; Top 10 Technical Key Project in Hunan Province 2023GK1010; Key Technologies R\&D Program of Guangdong Province (2023B1111030004 to FFH). The Funds of State Key Laboratory of Chemo and Biosensing, the National Supercomputing Center in Changsha (http://nscc.hnu.edu.cn/), Peng Cheng Lab, and Key Laboratory of High-Performance Distributed Ledger Technology, Digital Finance (Ministry of Education), and Hunan Research Center of the Basic Discipline for Cell Signaling.

\bibliographystyle{IEEEtran}
{
\bibliography{refs}
}

\end{document}